\documentclass[runningheads]{llncs}
\usepackage[T1]{fontenc}
\usepackage{graphicx}
\usepackage{amssymb}
\usepackage{pifont,multirow,array}
\usepackage[colorlinks,allcolors=blue]{hyperref}
\begin{document}
\title{FaRMamba: Frequency-based learning and Reconstruction aided Mamba for Medical Segmentation}
%
%
\author{
  Ze Rong\inst{1} \and
  Ziyue Zhao\inst{1} \and
  Zhaoxin Wang\inst{2} \and
  Lei Ma\inst{2}\textsuperscript{\textdagger}
}
\authorrunning{Z. Rong et al.}

\institute{
  School of Artificial Intelligence and Computer Science, Nantong University, China \and
  School of Information Science and Technology, Nantong University, China \\[1ex]
    \textsuperscript{\textdagger}* Corresponding author.
}
\maketitle              
\begin{abstract}
Accurate medical image segmentation remains challenging due to blurred lesion boundaries (LBA), loss of high-frequency details (LHD), and difficulty in modeling long-range anatomical structures (DC-LRSS). Vision Mamba employs one-dimensional causal state-space recurrence to efficiently model global dependencies, thereby substantially mitigating DC-LRSS. However, its patch tokenization and 1D serialization disrupt local pixel adjacency and impose a low-pass filtering effect, resulting in Local High-frequency Information Capture Deficiency (LHICD) and two-dimensional Spatial Structure Degradation (2D-SSD), which in turn exacerbate LBA and LHD. In this work, we propose FaRMamba, a novel extension that explicitly addresses LHICD and 2D-SSD through two complementary modules. A Multi-Scale Frequency Transform Module (MSFM) restores attenuated high-frequency cues by isolating and reconstructing multi-band spectra via wavelet, cosine, and Fourier transforms. A Self-Supervised Reconstruction Auxiliary Encoder (SSRAE) enforces pixel-level reconstruction on the shared Mamba encoder to recover full 2D spatial correlations, enhancing both fine textures and global context. Extensive evaluations on CAMUS echocardiography, MRI-based Mouse-cochlea, and Kvasir-Seg endoscopy demonstrate that FaRMamba consistently outperforms competitive CNN-Transformer hybrids and existing Mamba variants, delivering superior boundary accuracy, detail preservation, and global coherence without prohibitive computational overhead. This work provides a flexible frequency-aware framework for future segmentation models that directly mitigates core challenges in medical imaging.


\end{abstract}

\section{Introduction}
Medical image segmentation is critical for  in clinical applications such as tumor detection, organ recognition, and lesion localization, significantly enhancing diagnostic accuracy, optimizing treatment planning, and ensuring surgical safety \cite{minaee2022image}. Three key challenges persist  First, due to the inherently low contrast and noise interference of medical images and partial volume effects, gradient cues at lesion–tissue interfaces are severely attenuated, causing LBA\textit{(blurred boundaries) that impede precise delineation.} Second, partial volume averaging, low-pass filtering, \textit{and motion artifacts induce LHD (high-frequency detail loss) by obscuring small lesion textures."} Third, bias fields and local artifacts disrupts global contextual coherence, while localized models fail to capture distant anatomical dependencies, leading to DC-LRSS.

Convolutional neural networks (CNNs), epitomized by U-Net, exploit hierarchical convolution and skip connections to aggregate multi-scale edge and texture cues, mitigating LBA and LHD. However, their restricted receptive field hinder global context modeling, failing on DC-LRSS. In contrast, Vision Transformers (ViT) employ global self-attention to capture long-range dependencies and solving  DC-LRSS. Nevertheless, their patch-based tokenization undermines local adjacency and reduces sensitivity to fine boundaries and high-frequency textures. This trade-off leads to degraded performance on small or low-contrast lesions. Additionally, the quadratic complexity of self-attention incurs prohibitive computation.

The State-Space Model (SSM) \cite{gu2023modeling}, an effective architecture for long-sequence analysis, have attracted significant attention.  Mamba \cite{gu2024mamba}, an SSM variant, surpasses Transformers \cite{vaswani2017attention} in long-range dependency modeling. With the advent of ViM and Vmamba \cite{zhu2024vision}, Mamba has been successfully applied to computer vision gradually, achieving strong results. Medical adaptation UMamba  \cite{ma2024umamba} has demonstrated exceptional performance due to its efficient design, offering a promising solution for long-range dependency modeling in biomedical image analysis. Specifically, Vision Mamba–type models mitigate DC-LRSS by employing a structured state-space recurrence mechanism to perform global feature aggregation across whole targets, and their linear computational complexity yields substantially lower computational and memory overhead compared to ViT, enabling superior scalability to large input sizes and batch processing. However, two bottlenecks hinder Mamba’s medical application. First, the reliance on patch embedding and one-dimensional serialization disrupts local pixel adjacency and attenuates high-frequency detail propagation. This  LHICD hampers the network's sensitivity to fine textures and boundary cues, exacerbating LBA and LHD. Second, Mamba's causal state-space recurrence scans data as a 1D sequence, failing to fully preserve two-dimensional spatial correlations. This 2D-SSD undermines the model's ability to infer both local geometry and global anatomy simultaneously, leaving DC-LRSS insufficiently addressed. Addressing LHICD and 2D-SSD is therefore critical to realizing Mamba's full potential for precise, detail-preserving segmentation in complex biomedical imaging scenarios.

To tackle Mamba's core limitations (LHICD/2D-SSD) and mitigate LBA, LHD and DC-LRSS, we propose FaRMamba, a Vision Mamba variant integrating dual modules. The main contributions are as follows:
\begin{itemize}
	\item MSFM addresses LHICD via multi-scale frequency decomposition, enabling hierarchical feature extraction. We explore the feature processing capabilities of Discrete Wavelet Transform (DWT), Fast Fourier Transform (FFT), and Discrete Cosine Transform (DCT) within the same framework, providing a comprehensive analysis of their respective advantages in feature extraction.
	\item SSRAE resolves 2D-SSD through pixel-wise reconstruction, injecting spatially coherent priors to enhance joint local-global anatomy inference.
	\item FaRMamba outperforms highly competitive CNN–Transformer hybrids and Mamba variants on Mouse-cochlea, CAMUS \cite{leclerc2019deep}, and Kvasir-SEG\cite{jha2020kvasirseg}, delivering better boundary accuracy, detail preservation, and long-range structure modeling.
\end{itemize}

\section{Related Work}

\subsection{Vision Mamba}
In recent years, SSMs \cite{gu2023modeling}, particularly Structured State-Space Sequence (S4) models \cite{gu2022efficiently}, have emerged as efficient building blocks (or layers) for constructing deep networks, achieving SOTA performance in continuoussequence modeling. Mamba further enhances S4 \cite{gu2024mamba} by incorporating a selective mechanism, enabling the model to choose relevant information in an context-aware.Unlike Transformers, Mamba uses selective propagation guided by causal principles.Mamba has shows superior performance over CNNs and Transformers in computer vision (CV) tasks, making it a promising backbone for CV applications. As a result, considerable attention has recently been focused on applying Mamba to CV tasks \cite{xu2024survey}. Frequency-domain analysis is a key technique in signal processing that reveals the frequency components of a signal by transforming it from the time or spatial domain to the frequency domain. Some recent studies have explored the impact of frequency domain processing on Mamba. For example, GlobalMamba \cite{wang2024globalmamba} uses the DCT to divide images into different frequency domains and employs lightweight CNNs to process these frequency bands as causal sequences. EM-Net \cite{chang2024emnet} combines Fast Fourier Transform (FFT) with channel compression and adaptive calibration mechanisms to perform multi-scale feature extraction. P-Mamba \cite{ye2024pmamba} integrates Discrete Wavelet Transform (DWT) with Perona-Malik Diffusion (PMD) for noise suppression and boundary preservation.

\subsection{Reconstruction‐driven Feature Enhancement}
Auxiliary branches enhance segmentation by preserving fine textures, boundary cues, and global coherence in latent representations. Early evidence comes from Y-Net, which augments a U-shape backbone with a parallel decoder that reconstructs tissue appearance; this sharpens latent representations and yields higher breast-biopsy segmentation fidelity \cite{mehta2018ynet}. A photoacoustic variant of Y-Net shows that jointly optimising reconstruction and segmentation recovers of high-frequency details lost during tomographic inversion \cite{lan2020ynet}. Self-supervised masked-reconstruction objectives to large, unlabelled corpora. MedMAE masks up to 75 \% of voxels and trains a ViT encoder to predict the missing content; fine-tuning that encoder boosting Dice on organ segmentation benchmarks \cite{zhou2023self}, while context restoration report similar gains across CT and MR modalities \cite{chen2019selfsupervised}. Spatial-slice alignment and saliency prediction in SSL frameworks strengthen the reconstruction-segmentation synergy \cite{li2024selfsupervised}. joint reconstruction \& segmentation pipelines confirm the benefit under extreme artefacts: integrating a shape-aware reconstructor with a 3-D knee-MRI segmentor reduces cartilage errors versus two-stage baselines \cite{kofler2024joint}.

\section{Method}

\subsection{Architecture overview}
\begin{figure}[th!]
	\centering
	\includegraphics[width=1\linewidth]{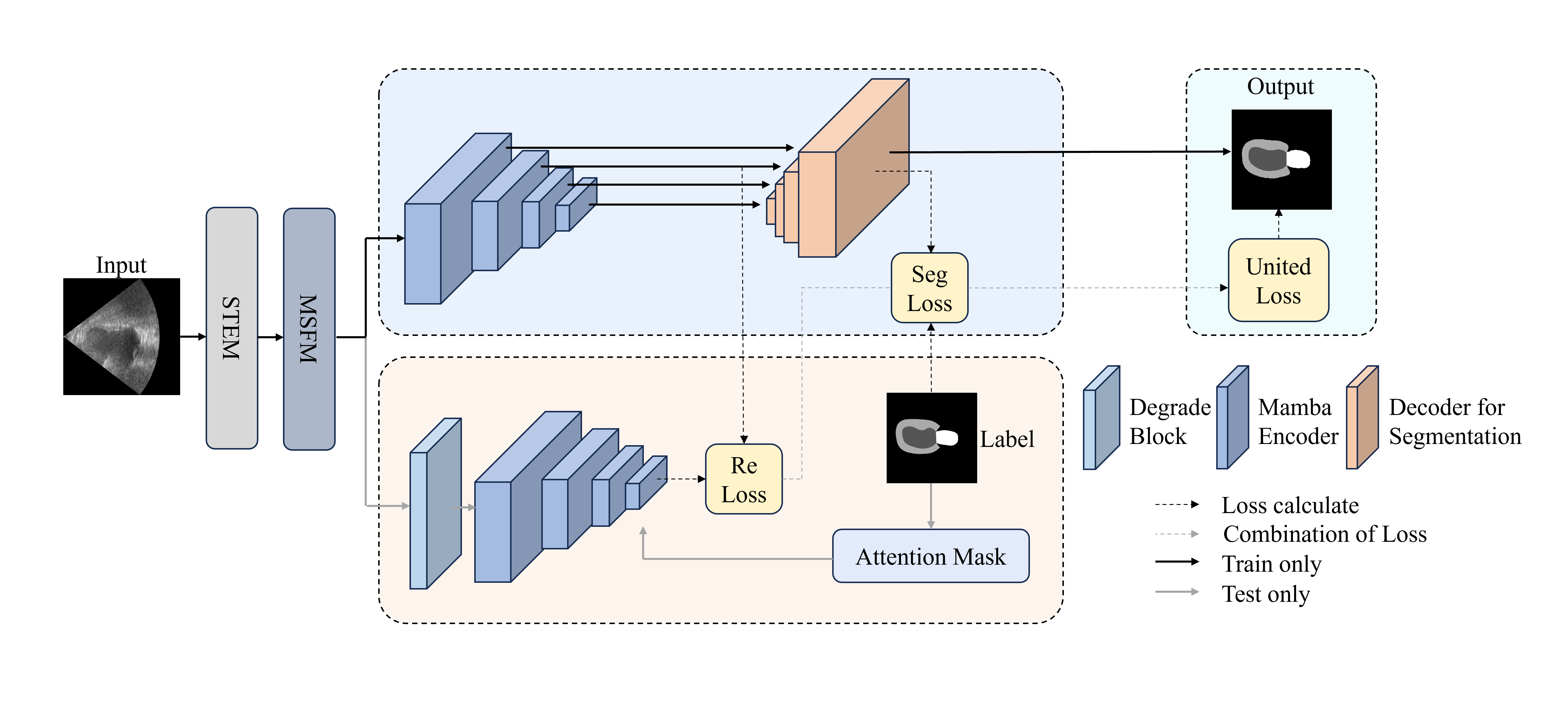}
	\caption{Architecture of FaRMamba.} 
	\label{fig:fig1}
\end{figure}

FaRMamba comprises four components: the primary encoder for the segmentation branch, SSRAE, MSFM, and the decoder. As shown in Fig. \ref{fig:fig1}, the STEM layer handles initial feature extraction and normalization. MSFM then leverages multi-scale frequency analysis. Concurrently, the primary segmentation branch encoder extracts image features while the SSRAE reconstructs segmentation encoder outputs from degraded inputs.Label-guided region attention in SSRAE prioritizes segmentation-identified ROIs to boost cross-task synergy. The model incorporates the decoder from UMamba \cite{ma2024umamba} and utilizes a joint loss function for backpropagation. 

\subsection{MSFM}
In the Mamba network, the modeling of input data relies on tokenization and recurrence mechanisms. First, patch embedding splits 2D feature map into fixed-size, non-overlapping patches and into fixed tokens, averaging pixel intensities when the patch size approaches lesion dimensions; high-frequency gradients and micro-textures are thus suppressed at the outset. Second, the one-dimensional state-space recurrence implements a stable linear dynamical system \cite{wang2023statespace} whose impulse response decays exponentially \cite{oppenheim2017signals}—a characteristic low-pass filter behavior that further attenuates any remaining high-frequency components during recurrence; residual high-frequency components are therefore further attenuated during recurrence. High-frequency boundary gradients and subtle textures are progressively attenuated by patch-level averaging followed by low-pass filtering in the state-space recurrence \cite{yu2024tuning}, giving rise to LHICD and thereby motivating the explicit restoration of multi-scale frequency cues via MSFM.

To counter high-frequency cues in Vision Mamba, MSFM's role is to project spatial feature maps into the frequency domain and aggregate information across multiple spectral bands for multi-scale learning. To this end, each intermediate feature map is subjected to a frequency-domain transformation method, producing a set of band-specific spectra. Frequency masks or sub-band filters are then applied to isolate distinct frequency ranges, producing masked spectra highlighting complementary frequencies content. These masked spectra are processed through scale-adapted convolutional blocks in the frequency domain to extract multiscale representations, and the outputs are subsequently inversely transformed to spatial features. The reconstructed feature maps are finally fused to augment the original spatial representation with restored high-frequency detail and enhanced boundary and texture sensitivity for the segmentation task. Since no single frequency transform universally outperforms others in biomedical imaging, Discrete Wavelet Transform (DWT), Discrete Cosine Transform (DCT), and Fast Fourier Transform (FFT)—are systematically evaluated in this study.

\subsubsection{DWT(Discrete Wavelet Transformer) }
First, Discrete Wavelet Transform (DWT) is applied to the image. The wavelet transform generates a series of wavelet basis functions $\psi_{j,k}(t)$ through the translation and scaling of the mother wavelet function  $\psi(t)$, which is used to represent the local features of the signal. The basis function is defined as: :
\begin{equation}
	\psi_{j,k}\left(t\right)=\frac{1}{\sqrt{2^j}}\psi\left(\frac{t-2^jk}{2^j}\right)
\end{equation}

Here, $\psi(t)$ represents the mother wavelet function, $j$ is the scale factor indicating the scaling degree of the wavelet function, and $k$ is the translation factor indicating the position of the wavelet function. Varying \textit{j} and \textit{k} yields multi-scale/spatial basis functions, capturing time-frequency-scale localized features. DWT coefficients \textit{W}\textit{j},\textit{k} are computed as: 
\begin{equation}
	W_{j,k}=\int_{-\infty}^{\infty}f\left(t\right)\psi_{j,k}\left(t\right)dt
\end{equation}

Here, $W_{j,k}$ represents the projection coefficient of the signal $f\left(t\right)$ onto the wavelet basis function  $\psi_{j,k}\left(t\right)$, reflecting the feature of the signal at scale $j$ and position $k$.

\begin{figure}[th!]
	\centering
	\includegraphics[width=1\linewidth]{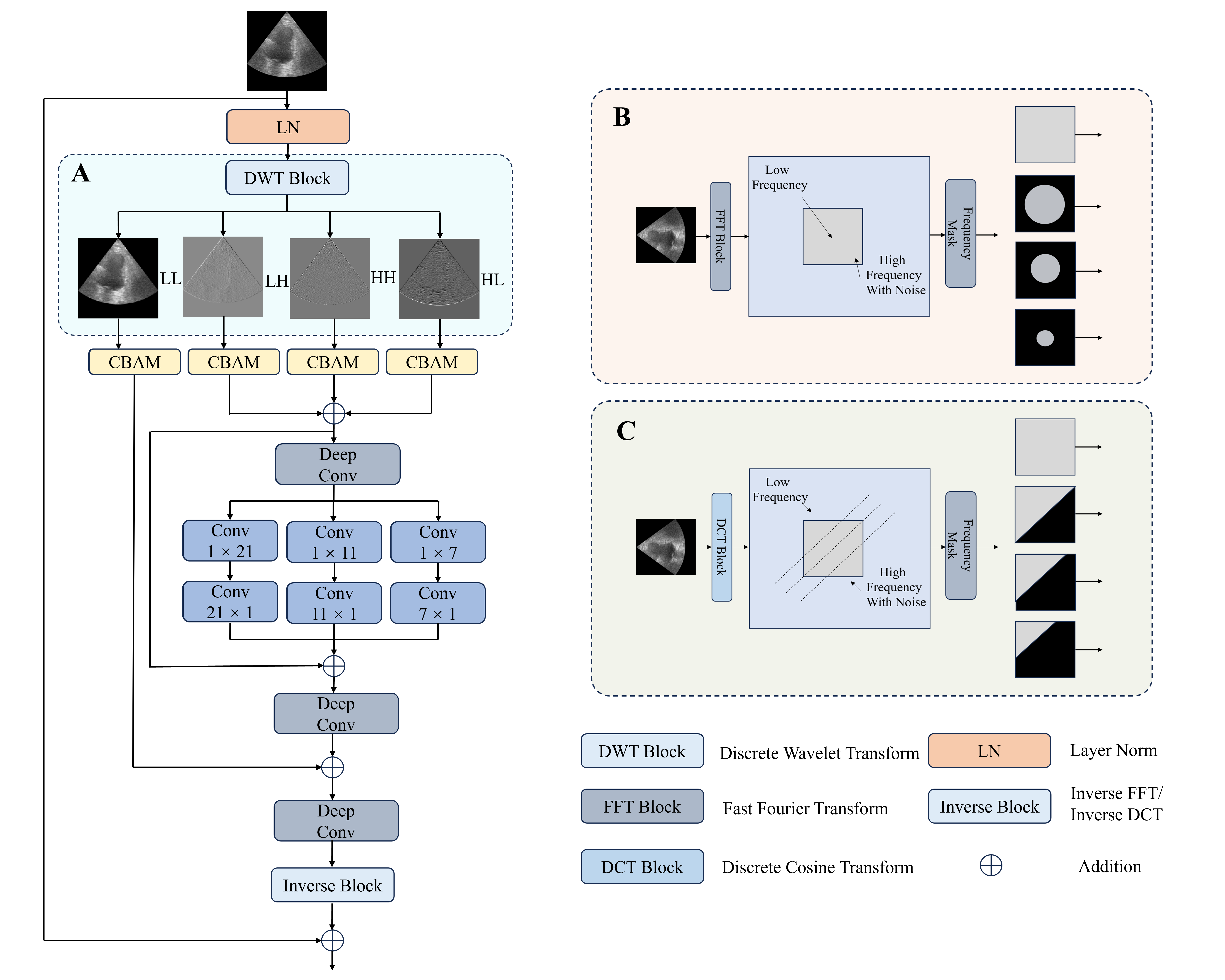}
	\caption{Architecture of MSFM.}
	\label{fig:fig2}
\end{figure}

We adopt the first-order Daubechies wavelet for balanced computational efficiency and time-frequency localization. The detailed structure of the wavelet transform module is shown in Fig. \ref{fig:fig2} First, the input tensor is split into four sub-bands: the low-frequency sub-band (LL) and the high-frequency sub-bands (LH, HL, HH), each containing information at different scales of the image. To adaptively recalibrate channel and spatial importance, the CBAM \cite{huang2024mambamir} module is applied to each sub-band, enabling the network to focus on both global and local features of the image simultaneously.Direct IDWT fusion of low/high-frequency features risks suboptimal detail preservation. Instead of directly using IDWT, we fuse the high-frequency sub-bands (LH, HL, HH) and apply a set of convolutional kernels with varying scales to enhance the features, further accentuating the high-frequency details. Finally, the fused high-frequency features are combined with the low-frequency sub-band to complete the process.

\subsubsection{FFT (Fast Fourier Transform)	}
The input image is first transformed to the frequency domain via FFT. The formula for FFT is as follows:
\begin{equation}
	F(u,v)=\sum_{x=0}^{M-1}\sum_{y=0}^{N-1}f(x,y)e^{-j2\pi(\frac{u_x}{M}+\frac{v_y}{N})}
\end{equation}
Where $f\left(x,y\right)$ represents the pixel values of the image in the time domain, and $F\left(u,v\right)$ denotes the transformed image in the frequency domain. $M$ and $N$ are the dimensions of the image, and $u$ and $v$ are the coordinates in the frequency domain.

FFT generates a complex spectrum: low frequencies cluster at the origin, high frequencies radiate outward.Concentric circular masks with decreasing radii isolate distinct frequency bands. Zeroing coefficients outside each mask yields band-specific spectra. Subsequently, each masked spectrum is subjected to inverse FFT (IFFT), yielding spatial-domain feature maps with frequency-specific information. These reconstructed maps are then fused into the main feature pipeline for downstream processing.

\subsubsection{DCT (Discrete Cosine Transform)}
Similarly, DCT is applied to the image processing. The formula for DCT is as follows:
\begin{equation}
	X(u,v)=\sum_{x=0}^{M-1}\sum_{y=0}^{N-1}f(x,y)cos\left[\frac{\pi(2x+1)u}{2M}\right]cos\left[\frac{\pi(2y+1)}{2N}\right]
\end{equation}
Where $f\left(x,y\right)$ represents the pixel values of the image in the time domain, and $X\left(u,v\right)$ is the transformed image in the frequency domain. $M$ and $N$ are the image dimensions, and $u$, $v$ are the frequency domain coordinates

DCT arranges low-frequency coefficients in the upper-left, high-frequency components and noise in the lower-right. Upper-triangular masks of increasing size enable fine-grained frequency control. Each successive mask penetrates further into the high frequency region. For each mask, coefficients outside the triangular region are zeros, isolating the target frequency band. Each masked DCT spectrum is then subjected to IDCT to reconstruct a spatial-domain feature map enriched with the selected frequency characteristics. These maps are subsequently reintegrated into the main feature pipeline for downstream processing.

\subsection{Segmentation Branch Main Encoder}
The main encoder of segmentation branch in our model is built using the VSS module \cite{zhu2024vision}, which is designed to effectively model long-range dependencies. The core computation unit of the VSS block is the 2D Selective Scanning (SS2D) method. SS2D extends image patches in four directions to create sequences, which are then processed individually by the Selective Scanning Model (SSM). Finally, the extracted features are combined into a complete 2D feature map.

To capture multi-scale cues, the frequency module is placed early in shallow layers, enabling cross-scale feature extraction for global modeling. As shown in Fig. \ref{fig:fig3}, after Patch Embedding, the image is divided into patches of size $4\times4$, resulting in a tensor size of $\mathbb{R}^{B\times96\times\frac{H}{4}\times\frac{W}{4}}$. This tensor is then processed by the SS2D module for feature modeling. SS2D processes this tensor, followed by patch merging (combining 4 patches) to produce $\mathbb{R}^{B\times192\times\frac{H}{8}\times\frac{W}{8}}$.Repeating this (SS2D + patch merging) yields the final tensor $\mathbb{R}^{B\times768\times\frac{H}{32}\times\frac{W}{32}}$.

\subsection{SSRAE}
In Vision Mamba, 2D spatial correlations are disrupted when feature maps are linearized into a unidirectional token sequence, as adjacent pixel relationships and global anatomical layouts become dispersed during causal recurrence. As a result, the model's capacity for preserving coherent spatial context and reconstructing geometric details is severely impaired, a phenomenon referred to as 2D-SSD. To overcome 2D-SSD, mechanisms are required that explicitly enforce both global positional consistency and local boundary fidelity. Self-supervised reconstruction objectives have been shown to encourage encoders to learn spatially coherent representations by requiring precise restoration of input structures—thereby enhancing downstream segmentation performance through improved boundary alignment and texture preservation \cite{li2024selfsupervised}. Motivated by these findings, a SSRAE is introduced to recover and inject structurally rich features back into the segmentation backbone.

SSRAE adopts the identical Mamba-based architecture used in the segmentation backbone to ensure that reconstruction-derived features reside in the same latent space as segmentation representations, thereby avoiding domain mismatches that can arise when fusing heterogeneous encoders \cite{chen2016semantic}. Moreover, parameter sharing across tasks reduces overall model complexity and training overhead compared to maintaining a separate reconstruction network \cite{caruana1997multitask,ruder2017overview}. Hence, SSRAE leverages the existing Mamba to recover and inject structurally coherent spatial features into the segmentation stream

\begin{figure}[th!]
	\centering
	\includegraphics[width=1\linewidth]{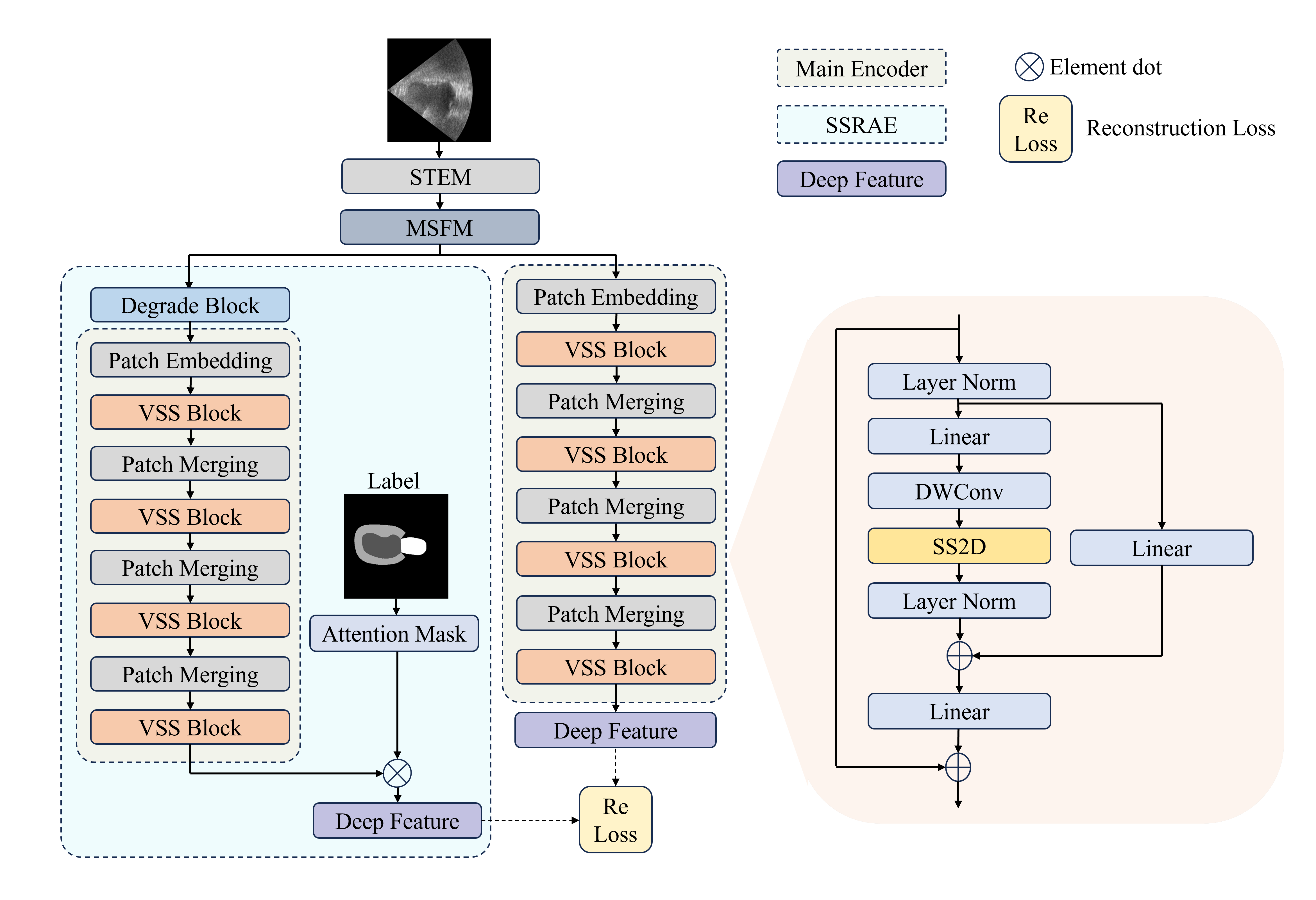}
	\caption{Architecture of the Segmentation Branch Main Encoder and SSRAE. }
	\label{fig:fig3}
\end{figure}

The detailed structure is shown in Fig. \ref{fig:fig3}, the image passes through the STEM layer into a Degrade Block that simulates acquisition noise/distortion. The Degrade Block simulates image quality degradation by performing three operations. First, it downscales the input tensor $\mathbb{R}^{B\times C\times H\times W}$ to $\mathbb{R}^{B\times C\times\frac{H}{4}\times\frac{W}{4}}$, reducing the resolution to simulate compression or downscaling. Next, a depthwise convolution with a $3\times3$ kernel is applied to introduce a blurring effect, mimicking the loss of fine details. Finally, Gaussian noise with a standard deviation $\sigma_{noise}$ (default 0.01) is added to the tensor, simulating sensor noise or transmission artifacts. These operations reduce image quality by lowering resolution, blurring, and introducing noise. This degradation process enables the network to learn how to restore high-quality images from low-quality ones during training. This self-supervised process trains the network to restore high-quality images and learn low-dimensional representations without extra labels. 

Vision Mamba's patch embedding accelerates computation via non-overlapping tokens but disrupts local spatial coherence and introduces background interference.SSRAE introduces data-driven Region Attention to mitigate these issues. Binary region masks are first derived from segmentation labels and then resized to match the feature map resolution. Flattened masks (avoiding per-region iteration) construct a unified attention mask, where large negative values nullify cross-region pixel affinities. During multi-head self-attention, this mask is added directly to the attention logits, thereby constraining attention to remain within each labeled region. Finally, the high-aafrequency feature maps from the main encoder are selected as reconstruction targets, ensuring that the network focuses on restoring clinically relevant details without incurring significant extra computation.

\subsection{Joint Loss Function}

\paragraph{Segmentation Loss. }
Since composite loss functions have proven robust across different tasks \cite{ma2021loss}, effectively assessing the overlap between predicted segmentation results and ground truth labels, the segmentation loss is designed as a weighted sum of Dice Loss and Cross-Entropy Loss.
\paragraph{Reconstruction Loss. }
Traditional losses (SSIM/Perceptual) focus on low-level features (pixels, color) but fail to capture high-dimensional high-frequency similarities. To accurately recover the complex, high‑frequency features in our deepest layers, We use a composite reconstruction loss—L1 for fine‑detail fidelity, cosine similarity for semantic alignment, and gradient loss for edge and texture recovery—to directly constrain high‑frequency feature restoration, unlike SSIM or perceptual losses that focus on low‑level pixel and color discrepancies.

\paragraph{Joint Loss Function. }
Given that the reconstruction loss relies on the output of the segmentation encoder, a strategy of gradually introducing the reconstruction loss with dynamic weighting is adopted. In the initial stages of training, only the segmentation task is optimized, while the reconstruction loss is progressively introduced. The weight of the reconstruction loss is dynamically adjusted according to the training epoch, following a linear decay strategy. Additionally, the reconstruction loss weight is smoothed using Exponential Moving Average (EMA) to ensure the stability of the training process.

\section{Experiments}

\subsection{Datasets and Implementation Details}
We validate the performance and scalability of FaRMamba under various segmentation targets and settings on three different datasets:

\textbf{Mouse-cochlea} Private MRI dataset (583 images, 13 mice) from xxUniversity, annotating cochlear lymphatic regions, elliptical sac, and bone structures. Not publicly available due to privacy.

\textbf{CAMUS} : Public cardiac ultrasound dataset (500 patients, 2000 images) from GE Vivid E95, with A2C/A4C views. Annotates LV, LA, MYO. Evaluated via 10-fold CV. 

\textbf{Kvasir-SEG }: Public polyp dataset (1000 images) with gastroenterologist-annotated masks. Original resolutions vary (332×487–1920×1072); resized to 256×256. 

Implementation: PyTorch on Ubuntu 22.04/CUDA 11.8, single RTX 3090 (24GB VRAM), i9-12900K. Adam optimizer (lr=1e-4, batch=25), 100 epochs (validate every 10). Metrics: DSC and MIoU. 

\subsection{Performance Comparison}
Comparison experiments were conducted under identical training protocols between FaRMamba and representative CNN-based, Transformer-based, and frequency-based enhanced baselines. Quantitative results in Table \ref{tab1} indicate that FaRMamba outperforms other segmentation networks in medical image segmentation tasks. Among the three MSFM instantiations, a modality-dependent ranking is observed: FaRMamba-DWT attains the highest Dice on the CAMUS echocardiography set; FaRMamba-FFT leads on the MRI-based Mice-cochlea set; and FaRMamba-DCT achieves the best performance on the Kvasir-Seg endoscopic set.

\begin{table}[!ht]
	\centering
	\caption{Comparison with state-of-the-art methods.}
	\label{tab1}
	\extrarowheight2pt
	\tabcolsep4pt
	\begin{tabular}{lllllllll}
		\hline
		\multirow{2}{*}{Methods} & \multicolumn{2}{c}{MOUSE}  & \multicolumn{2}{c}{CAMUS(2CH)} & \multicolumn{2}{c}{CAMUS(4CH)} & \multicolumn{2}{c}{Kvasir-Seg} \\
		\cline{2-9}
		& DSC & MIoU & DSC & MIoU & DSC & MIoU & DSC & MIoU \\
		\hline
		ResNet18\cite{he2016deep} & 57.39 & 45.86 & 87.69 & 78.55 & 94.03 & 91.49 & 74.48 & 64.89 \\
		nnUNet\cite{isensee2021nnunet} & 59.82 & 48.49 & 88.50 & 79.79 & 93.62 & 88.24 & 82.34 & 74.42 \\
		UKAN\cite{li2025ukan} & 58.41 & 47.09 & 88.67 & 80.11 & 95.53 & \textbf{93.52} & 77.39 & 69.42 \\
		UNetFormer\cite{wang2022unetformer} & 58.42 & 46.78 & 86.85 & 77.39 & 94.56 & 90.34 & 63.80 & 54.94 \\
		TransUNet\cite{chen2021transunet} & 59.53 & 47.91 & 84.70 & 74.18 & 94.81 & 90.49 & 64.01 & 53.88 \\
		SwinUNet\cite{cao2023swinunet} & 57.99 & 46.56 & 85.19 & 75.01 & 94.47 & 89.89 & 63.64 & 52.29 \\
		WaveletUNet\cite{li2023globalfrequencydomain} & 58.45 & 46.35 & 88.36 & 80.12 & 95.75 & 92.05 & 73.22 & 64.13 \\
		GFUNet\cite{li2020wavelet} & 57.53 & 45.47 & 87.91 & 79.01 & 95.61 & 91.81 & 71.97 & 61.84 \\
		Umamba\cite{ma2024umamba} & 57.93 & 46.85 & 88.62 & 79.96 & 93.63 & 88.23 & 79.60 & 71.76 \\
		Ours-FFT & \textbf{60.89} & \textbf{48.91} & 89.39 & 81.25 & 95.56 & 91.78 &87.39 &  80.31 \\
		Ours-DCT & 59.98 & 48.68 & 89.60 & 81.50 & 95.86 & 92.37 & \textbf{88.97 } & \textbf{81.87} \\
		Ours-DWT & 60.13 & 48.69 & \textbf{89.81} & \textbf{81.88} & \textbf{95.98} & 92.57 & 87.86 & 80.55 \\
		\hline
	\end{tabular}%
\end{table}

These findings reflect known modality‐specific frequency characteristics and the complementary strengths of each transform. Speckle noise in echocardiography (CAMUS) \cite{nes2011fast,gupta2004waveletbased} obscures fine details; DWT’s spatial-frequency localization enhances edges and reduces noise, justifying its superiority.  In contrast, MRI inherently samples k-space (the Fourier domain) so global spectral analysis via the FFT naturally captures long-range correlations and faithfully reflects the physics of MR signal encoding, accounting for FFT's leading accuracy on the Mice-cochlea dataset \cite{pruessmann1999sense,gosche2024domain}.Endoscopic imagery features rich mid-frequency textures and repetitive patterns; the discrete cosine transform's energy-compaction property concentrates key structural information into a small number of coefficients while suppressing high-frequency noise, enabling DCT's top performance on Kvasir-Seg \cite{ahmed1974discrete,zhao2022discrete}. Thus, selecting transforms based on modality noise/frequency profiles is theoretically sound and empirically validated, confirming MSFM’s integration of DWT/FFT/DCT. 

\begin{figure}[th!]
	\centering
	\includegraphics[width=1\linewidth]{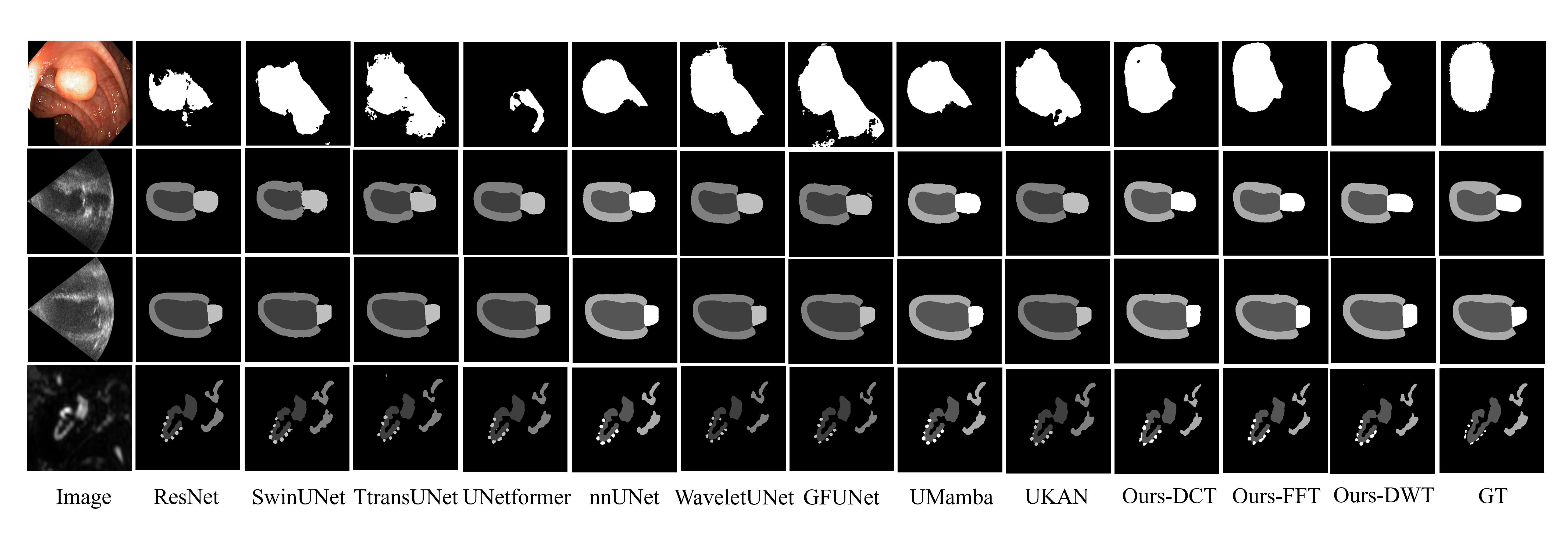}
	\caption{Qualitative Comparison of Segmentation Results Across Different Methods. }
	\label{fig:fig4}
\end{figure}

\subsection{Ablation Study}
An ablation study was conducted on the CAMUS dataset to assess the individual and combined effects of MSFM and SSRAE within the FaRMamba framework (Table \ref{tab2}). Six model variants were evaluated: Base (segmentation encoder only), Base + MSFM (segmentation branch), Base + SSRAE without MSFM in the reconstruction branch, Base + SSRAE with MSFM in the reconstruction branch, Base + MSFM + SSRAE without MSFM in the reconstruction branch, and the full FaRMamba configuration (MSFM in both branches). Integrating MSFM into the segmentation encoder resulted in a notable improvement in performance metrics, confirming its efficacy in restoring high-frequency information. Introducing SSRAE alone yielded additional gains by enhancing spatial coherence via reconstruction. When MSFM was also incorporated into the reconstruction branch, further performance enhancements were observed, underscoring the advantage of frequency-aware reconstruction. Finally, the complete FaRMamba configuration—featuring MSFM in both branches alongside SSRAE—delivered the highest overall performance, demonstrating that the two modules complementarily address LHICD and 2D-SSD and collectively achieve the most substantial improvements in segmentation accuracy.

\begin{table}[!ht]
	\centering
	\caption{Ablation study on CAMUS(2CH).}
	\label{tab2}
	\extrarowheight2pt
	\tabcolsep5pt
	\begin{tabular}{llllllllll}
		\hline
		\multirow{2}{*}{Main} & \multirow{2}{*}{MSFM} & \multirow{2}{*}{SSRAE} & \multirow{2}{7em}{SSRAE without MSFM} & \multicolumn{2}{c}{DWT} & \multicolumn{2}{c}{FFT}& \multicolumn{2}{c}{DCT}\\
		\cline{5-10}
		&   &   &  & DSC & MIoU  & DSC & MIoU 
& DSC &MIoU 
\\
		\hline
		\ding{51} &   &   &   & 87.25 & 79.89  & 87.25 & 79.89 
& 87.25 &79.89 
\\
		\ding{51} & \ding{51} &   &   & 88.60 & 80.88  & 88.33 & 80.27 
& 88.40 &80.22 
\\
		\ding{51} &   & \ding{51} &   & 88.25 & 80.59  & 88.35 & 80.20 
& 88.22 &79.99 
\\
		\ding{51} &   &   & \ding{51} & 87.98 & 80.52  & 87.98 & 80.52 
& 87.98 &80.52 
\\
		\ding{51} & \ding{51} &   & \ding{51} & 89.13 & 81.02  & 88.95 & 80.66 
& 88.55 &80.70 
\\
		\hline
		\ding{51} & \ding{51} & \ding{51} &   & \textbf{89.81} & \textbf{81.88}  & \textbf{89.39} & \textbf{81.25} & \textbf{89.60} &\textbf{81.50} \\
		\hline
	\end{tabular}
\end{table}

\section{Conclusion} 
In medical image segmentation tasks, precision is hindered by LBA, LHD, and DC-LRSS. Vision Mamba's linear-complexity state-space recurrence alleviates global dependencies and thus substantially through DC-LRSS, yet its patch tokenization and one-dimensional serialization introduce LHICD and 2D-SSD. To address these shortcomings, FaRMamba incorporates MSFM to explicitly restore attenuated high-frequency cues and SSRAE to reinstate full two-dimensional spatial coherence. By preserving Mamba's efficient global modeling while resolving LHICD and 2D-SSD, FaRMamba delivers a unified solution to the core challenges of LBA, LHD, and DC-LRSS. Extensive experiments on three diverse datasets (CAMUS ultrasound, MRI-based Mice-cochlea, and Kvasir-Seg endoscopy) demonstrate that FaRMamba consistently outperforms state-of-the-art CNN–Transformer hybrids and existing Mamba variants. Modality-specific instantiations of MSFM reveal that DWT excels on noisy ultrasound, FFT aligns best with MRI's native k-space structure, and DCT is most effective for textured endoscopic imagery—validating the flexibility of the multi-transform design. Ablation studies confirm that MSFM and SSRAE each contribute significant Dice gains and yield synergistic improvements when combined.

\begin{credits}
	\subsubsection{\ackname} We gratefully acknowledge the assistance of all those who have contributed to this research.

\end{credits}

\bibliographystyle{splncs04}
 \bibliography{references}

\end{document}